\documentclass{article}

\usepackage{iclr2027/iclr2027_conference,times}

\usepackage{graphicx}
\usepackage{booktabs}
\usepackage{xcolor}
\usepackage{amsmath}
\usepackage{amssymb}
\usepackage{algorithm}
\usepackage{algpseudocode}
\usepackage{placeins}
\usepackage[accsupp]{axessibility}
\usepackage{hyperref}
\usepackage{url}

\let\cite\citep

\hypersetup{
  pdftitle={LongVU-TTT: Causal Test-Time Training for Visual Resampling in Long Video Understanding},
  pdfauthor={Mahmoud Ahmed, Sameh Abdulah, Olatunji Ruwase, Sam Ade Jacobs, Mathis Bode, Mohamed Elhoseiny}
}

\begin{document}

\title{LongVU-TTT: Causal Test-Time Training for\\Visual Resampling in\\Long Video Understanding}

\author{
\begin{minipage}{\dimexpr\textwidth-2\tabcolsep\relax}
\centering
\textbf{Mahmoud Ahmed\textsuperscript{1}, Sameh Abdulah\textsuperscript{1},
Olatunji Ruwase\textsuperscript{2},}\\
\textbf{Sam Ade Jacobs\textsuperscript{3}, Mathis Bode\textsuperscript{4},
Mohamed Elhoseiny\textsuperscript{1}}\\[0.7ex]
{\normalfont
\textsuperscript{1}KAUST \quad
\textsuperscript{2}Snowflake \quad
\textsuperscript{3}Microsoft Inc.\\[-0.1ex]
\textsuperscript{4}J\"ulich Supercomputing Centre, Forschungszentrum J\"ulich}
\end{minipage}
}

% Show authors and suppress review line numbering. The venue header is replaced
% below with a neutral preprint label.
\iclrfinalcopy

\maketitle
\lhead{Preprint}

\begin{abstract}
Long-video MLLMs must model temporal change before a limited visual-token budget removes
most frame evidence. We introduce \textbf{LongVU-TTT}, which inserts a convolutional
Test-Time Training (TTT) resampler with causal fast-weight updates between the vision
encoder and the LLM.
Its grouped 2D fast weights adapt to each video and contextualize frame features before
compression, while a hybrid uniform-and-change-aware selector retains explicit visual
evidence for downstream reasoning. Under controlled conditions, TTT-Conv improves over
TTT-MLP by up to +2.12 and bidirectional Mamba2 by up to +3.04 on MLVU, and it is
stronger than attention- and fixed-state recurrent resamplers across three benchmarks.
Analysis shows that the fast weights behave as a temporal aggregation state rather than a
reliable long-horizon episodic memory: their benefit attenuates as evidence becomes more
distant, motivating explicit frame retention. LongVU-TTT processes up to 512 frames before
reducing them to 128 LLM frames and achieves competitive performance across five video
understanding benchmarks. Code and models will be publicly released.
\end{abstract}

\section{Introduction}

Recent advancements in video MLLMs~\cite{lin2023videollava,damonlpsg2023videollama} have
improved AI reasoning over temporal content. These models encode individual frames
independently via a pretrained vision encoder before feeding tokens into an
LLM~\cite{zhu2023minigpt,li2024llavaonevisioneasyvisualtask}, forcing the LLM to perform
temporal modeling within a constrained context window. As video lengths increase, token
counts grow linearly while quadratic attention makes long sequences prohibitively
expensive.

\noindent Existing efforts reduce spatiotemporal redundancy through spatial pooling, token
merging, or differential token dropping~\cite{bolya2022token,
li2025videochatflashhierarchicalcompressionlongcontext,yao2025timechat,choudhury2024don},
but generally compress frame-level features before explicitly modeling what changed.
Recurrent architectures such as State-Space Models~\cite{jiang2025storm,islam2025bimba}
offer temporal propagation before compression, but their flattened sequence operators do
not explicitly impose the 2D local-neighborhood interactions that complement the global
representations of the ViT encoder.

\noindent We introduce \textbf{LongVU-TTT}, which inserts a convolutional Test-Time Training
(TTT) resampler between the vision encoder and the LLM. It processes frames recurrently
through continuously updated grouped-convolutional fast weights, preserving the visual
feature grid while adapting to temporal change. The inner-loop gradient signals provide a
change-aware importance measure that, combined with uniform sampling, reduces sequences
of up to 512 frames to 128 LLM frames without an auxiliary learned scorer. Crucially, we
find that the fast state is best understood as a temporal aggregator rather than an
episodic store: selected frames retain the explicit evidence required for long-range
reasoning. Here, \emph{causal} refers specifically to the fast-weight update rule: each
frame is contextualized only by preceding updates. Importance normalization and the final
fixed-budget frame selection are performed after the full video has been processed.

\noindent Through a controlled comparison with a temporal Transformer, Gated DeltaNet, Mamba2,
and TTT-MLP, we study which projector-stage temporal operator is most effective.
Our results identify causal update ordering and explicit 2D local processing as important
design choices, while retention diagnostics clarify the boundary between temporal
aggregation and long-horizon memorization. Our goal is not to claim a new state of the
art across models trained with heterogeneous data and backbones, but to provide a
controlled study of TTT at the projector stage of a video MLLM. We therefore emphasize
comparisons with models of similar scale that start from the same trained baseline.
System optimizations enabling 512-frame training on A100 40GB GPUs are detailed in
Section~\ref{supp:system} of the supplementary material.

\noindent Our contributions are as follows:
\begin{itemize}
    \item A convolutional TTT resampler with causal fast-weight updates that adapts
    grouped 2D fast weights to
    contextualize video features before the LLM's visual-token bottleneck.

    \item A saliency-driven hybrid frame selector that combines gradient-direction shift
    and alignment-loss delta to retain scene transitions and within-scene visual
    change without an auxiliary learned scorer.

    \item A controlled study spanning temporal Transformers, gated linear recurrence,
    Mamba2, and TTT variants, together with retention diagnostics showing that projector
    fast weights provide temporal aggregation rather than reliable episodic storage.
\end{itemize}

\section{Related Work}

\subsection{Video Multimodal Large Language Models}

The prevailing paradigm in video MLLMs involves treating videos as sequences of 
independent frame embeddings extracted via pretrained vision encoders such as 
CLIP~\cite{sun2023eva} or SigLIP~\cite{tschannen2025siglip}, and projecting them into 
the input space of an LLM~\cite{li2024llavaonevisioneasyvisualtask,damonlpsg2023videollama}. 
While effective for short-range spatial reasoning, these models delegate temporal modeling 
entirely to the LLM's self-attention mechanism, which becomes computationally prohibitive as video length increases.

\noindent To manage the resulting token count, various compression strategies have been proposed. 
Early methods employ uniform temporal or spatial pooling to reduce the visual 
footprint~\cite{wang2024qwen2vlenhancingvisionlanguagemodels}, while more recent adaptive 
approaches such as token merging and differential token 
dropping~\cite{bolya2022token,li2025videochatflashhierarchicalcompressionlongcontext,
yao2025timechat,choudhury2024don} perform more selective reduction. LongVU~\cite{shen2024longvuspatiotemporaladaptivecompression} 
takes a more structured approach by leveraging DINOv2~\cite{oquab2024dinov2learningrobustvisual} 
features to identify and remove redundant frames, and further applies text-guided 
cross-modal queries to selectively reduce frame features based on the input prompt. 
However, all of these methods operate on raw frame-level features without first 
propagating temporal context, and therefore compress without a principled understanding 
of what has changed across frames. In contrast, LongVU-TTT derives its compression 
signal directly from the TTT inner-loop updates, capturing visual change as 
a natural byproduct of temporal propagation rather than relying on external similarity 
metrics or auxiliary models.

\subsection{Recurrent Models and Test-Time Training}

Linear-complexity recurrent models offer an alternative to the quadratic cost of 
Transformer self-attention for long sequence 
modeling~\cite{vaswani2017attention,dao2024transformers}. Methods such as 
VideoMamba~\cite{li2024videomamba} and STORM~\cite{jiang2025storm} apply selective scan 
operations over video tokens to capture temporal dependencies prior to compression. 
While effective, these methods apply flattened sequence operators that do not explicitly
encode the 2D local-neighborhood interactions useful for fine-grained visual understanding.
Gated DeltaNet~\cite{yang2024gated} augments efficient linear recurrence with a gated
delta-rule update; we include it as a fixed-state recurrent comparison at the same
projector location as our method.

\noindent Test-Time Training (TTT)~\cite{sun2024learning} replaces fixed recurrent hidden states 
with fast-weight~\cite{schlag2021linear} models that are updated as a function of the 
input sequence, acting as a continuously evolving adaptation state. Recent TTT variants such as 
Titans~\cite{behrouz2024titans} and Atlas~\cite{behrouz2025atlas} have improved the 
memory capacity of these fast-weight models, but remain focused on 1D sequence modeling 
with MLP-based fast weights. LaCT~\cite{zhang2025test} improves the hardware efficiency 
of TTT-MLP by maximizing chunk size and head dimension to better utilize GPU 
tensor cores, and serves as the basis for our TTT-MLP baseline. We depart from this 
convention by parameterizing the fast weights as grouped convolutions, preserving the 
2D spatial structure of frame features during temporal propagation and capturing local 
spatial patterns that are complementary to the global representations of the ViT encoder. 
Rather than changing the LLM architecture itself, we place the temporal operator at the
projector stage of a pretrained MLLM and fine-tune the complete model end to end. This
setting lets us identify the architectural choices---including causal fast-weight updates
and preservation of spatial structure---that make projector-stage TTT effective.

\subsection{Efficient Training for Long-Video Models}

Training on long video sequences creates activation bottlenecks that parameter-sharding
methods such as DeepSpeed ZeRO-3~\cite{rajbhandari2020zero} do not remove. Sequence
parallelism~\cite{jacobs2023deepspeed}, ring attention~\cite{liu2023ring}, and systems
such as LongVILA's MM-SP~\cite{chen2024longvila} distribute long attention sequences,
but require specialized communication across devices. A complementary direction is
operator tiling: Arctic Long Sequence Training (ALST)~\cite{bekman2025arctic} partitions
LLM MLP computation into tiles and defers ZeRO-3 gradient synchronization until the
final tile. We adapt this idea to the independently processed ViT frame batch and pair it
with asynchronous activation-memory management; the method and measurements are given
in Section~\ref{sec:system-optimizations}.

\section{Methodology}

\subsection{Architecture Overview}

\begin{figure*}[t]
    \centering
    \includegraphics[width=\textwidth]{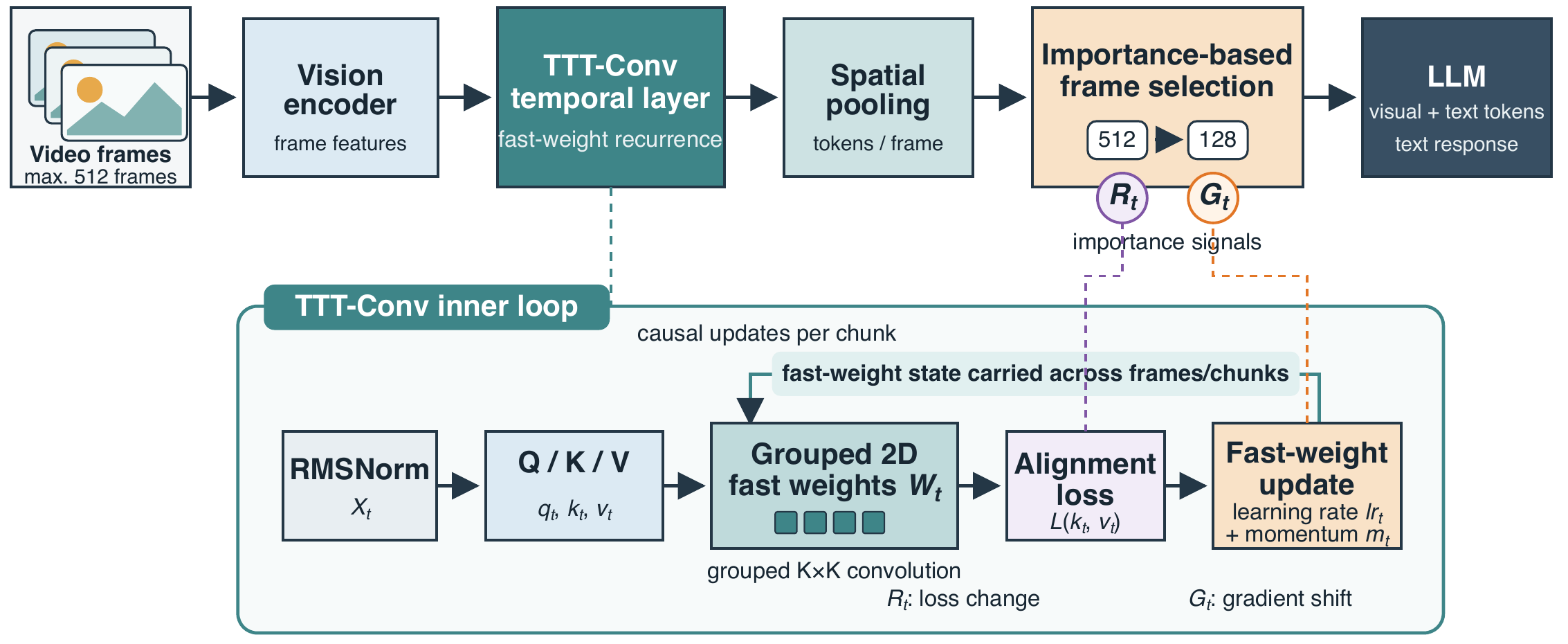}
    \caption{Overview of LongVU-TTT. Video frames are encoded independently by a 
    pretrained vision encoder to produce raw visual tokens. The TTT-Conv layer then 
    processes all frames recurrently via continuously updated grouped convolutional fast 
    weights, propagating temporal context across the sequence in chunks. The temporally 
    enriched representations are passed through spatial pooling and importance-based frame 
    sampling to produce a compact set of visual tokens. The existing MLP projector maps
    the selected features into the LLM embedding space, where they are concatenated with
    text embeddings and used to generate the response.}
    \label{fig:architecture}
    \vspace{-5pt}
\end{figure*}

LongVU-TTT introduces a dedicated temporal processing stage between the Vision 
Transformer (ViT) encoder and the Large Language Model (LLM), as illustrated in 
Figure~\ref{fig:architecture}. Rather than delegating all temporal reasoning to the LLM, 
this stage processes frames recurrently before they reach the LLM, enriching each frame's 
representation with context from all preceding frames and enabling principled compression 
of long video sequences. The layer receives frame-level features 
$X \in \mathbb{R}^{T \times (H \cdot W) \times C}$ from the ViT, where each frame 
maintains a spatial resolution of $H \times W = 27 \times 27$ tokens. To stabilize 
optimization, the TTT-Conv stage is wrapped in a residual block preceded by an RMSNorm 
layer:
\begin{equation}
X_{out} = X + \text{TTT-Conv}(\text{RMSNorm}(X))
\end{equation}
After spatial pooling and frame selection, the existing MLP projector maps the selected
features from the vision-encoder width to the LLM embedding width. Thus, TTT-Conv is a
projector-stage temporal operator, while the MLP projector remains the modality-alignment
mapping used by the base architecture.
The following sections describe the parameterization of the TTT-Conv layer, its 
inner-loop update rule, the frame selection strategy it enables, and the system 
optimizations required for scalable long-video training.

\subsection{TTT-Conv Parameterization and Projections}

The key design choice that distinguishes our TTT layer from prior work is its 
convolutional fast-weight parameterization, which directly leads to consistent 
improvements of up to $+2.12$ on MLVU over TTT-MLP and up to $+3.04$ over 
bidirectional Mamba2 under controlled conditions. Rather than flattening frame features 
into 1D sequences as TTT-MLP and SSM methods do, we parameterize the hidden state 
as two grouped convolutional kernels that operate directly on the 2D spatial structure 
of each frame. We use $N_g$ groups to partition the channel dimension, allowing each 
group to specialize in a subset of feature channels while maintaining parameter 
efficiency. Let $d=C/N_g$ be the input channels per group and $D_{int}=4d$ the
per-group intermediate width. The two fast-weight tensors are
$W_1\in\mathbb{R}^{(N_gD_{int})\times d\times K\times K}$ and
$W_2\in\mathbb{R}^{C\times D_{int}\times K\times K}$, which project each group into
and out of its intermediate representation. Frame-wise linear heads produce keys and
values $(k_t,v_t)$ for the inner update, queries $q_t$ for retrieval, two per-group
learning rates $lr_t\in(0,1)^{N_g\times2}$, and per-group momentum coefficients
$m_t\in(0,1)^{N_g}$. Both gates use a sigmoid; zero-initialized learning-rate logits
therefore begin at $\operatorname{sigmoid}(0)=0.5$.

\subsection{The TTT-Conv Inner Loop}

The TTT hidden state is a function rather than a fixed vector. Specifically, the fast
weights $(W_{1,t}, W_{2,t})$ define a convolutional mapping
$f_{W_t}: \mathbb{R}^{H \times W \times C} \rightarrow \mathbb{R}^{H \times W \times C}$ 
that is updated as frames arrive. It contextualizes the current frame from the observed
video history, but is not assumed to store individual past frames.
Frames are processed in chunks of fixed size $C_{size}$, with the weight state carried 
over across chunk boundaries to maintain temporal continuity across the full video.
The projections and initial fast weights are learned by the outer training objective.
At inference, the fast weights and momentum reset to their learned initial state at each
video boundary and adapt using only the self-supervised alignment objective, without
answer labels.

\noindent At each frame $t$, the fast weights are updated by minimizing a self-supervised 
alignment objective. Given the key $k_t$ and value $v_t$ projections of the current 
frame, the inner loop optimizes:
\begin{equation}
\mathcal{L}(k_t, v_t) = -\langle f_{W}(k_t),\, v_t \rangle
\end{equation}
which encourages the mapping to produce representations aligned with the current frame's
value. Empirically, frames that introduce visual change produce larger or directionally
different updates, a property later used for importance-based frame selection. %This property is later exploited for importance-based frame 
%selection.

\subsubsection{Parallel Causal Materialization}

A purely sequential inner-loop update preserves causal ordering but limits GPU
utilization. Conversely, treating each chunk as a single minibatch removes distinct
intra-chunk states. We retain causal ordering while parallelizing the computation.

\noindent We resolve this with a Parallel Causal Materialization strategy. For each frame $t$ 
within the chunk, we first compute its raw gradient independently:
\begin{equation}
g_t = -\nabla_{W} \mathcal{L}(k_t, v_t)
\end{equation}
Per-group learning rates first produce the scaled gradient $u_t=lr_t\cdot g_t$. Given
the carried momentum state $p_{init}$ from the previous chunk, the pre-update history
$h_t$ and
momentum-adjusted fast-weight increment $\delta_t$ are
\begin{equation}
h_t = p_{init}+\sum_{i=1}^{t-1}u_i, \qquad \delta_t=u_t+m_t\cdot h_t.
\end{equation}
The causal weight state excludes the current frame's own update:
\begin{equation}
W_t = W_{init}+\sum_{i=1}^{t-1}\delta_i.
\end{equation}
Here $g_t$, $u_t$, and $\delta_t$ denote the raw gradient, learning-rate-scaled gradient,
and momentum-adjusted increment, respectively. The coefficient $m_t \in (0,1)$ controls
the contribution of accumulated update history. At a chunk boundary, the final
momentum-adjusted increment is carried as $p_{init}$ for the next chunk. Using
\texttt{torch.vmap} and prefix sums, all per-frame weight states within a chunk are
materialized simultaneously. This trades activation memory for parallel compute; we use
asynchronous activation checkpointing to rematerialize these states during the backward
pass instead of retaining the full materialized tensor. Implementation details are in
Section~\ref{supp:system}.

\subsubsection{Parallel Apply Operation}

Following materialization, the query $q_t$ is transformed through its corresponding 
weight state $(W_{1,t}, W_{2,t})$ in parallel across all frames in the chunk:
\begin{equation}
z_t = \sigma\!\left(\text{Conv2d}(q_t, W_{1,t})\right), \qquad 
\hat{x}_t = \text{Conv2d}(z_t, W_{2,t})
\end{equation}
where $\sigma$ denotes the SiLU activation. The output $\hat{x}_t$ is a temporally 
contextualized representation of frame $t$ influenced by preceding fast-weight updates
before being passed to the LLM.

\subsection{Saliency-Driven Hybrid Frame Sampling}
\label{sec:importance}

A key property of our TTT layer is that the gradient signals produced during the 
inner-loop optimization provide a natural and principled measure of frame importance, 
requiring no additional learned components. This allows us to compress sequences of up 
to 512 frames to 128 frames while retaining the most informative content, as validated 
by the $+6.4$ MLVU gain over a baseline that subsamples without temporal propagation.

\subsubsection{Importance Scoring}

We extract two complementary scalar signals from the TTT inner loop to identify salient 
frames, where $t \in \{1, \ldots, T\}$ denotes the frame index.

\noindent \textbf{Gradient-Direction Shift ($G_t$)} captures scene-level changes by measuring 
the cosine distance between the fast-weight gradients of successive frames, yielding a 
scalar in $[0, 2]$ that is large when the direction of the fast-weight update shifts 
abruptly between frames $t-1$ and $t$:
\begin{equation}
G_t = 1 - \frac{\nabla_{W} \mathcal{L}_t \cdot \nabla_{W} \mathcal{L}_{t-1}}
{\|\nabla_{W} \mathcal{L}_t\| \|\nabla_{W} \mathcal{L}_{t-1}\|}
\end{equation}

\noindent \textbf{Alignment-Loss Delta ($R_t$)} captures event-level changes within a scene 
by measuring positive increases in the per-frame inner-loop alignment loss between 
consecutive frames, yielding a non-negative scalar that reflects how much less well
aligned frame $t$ is than frame $t-1$:
\begin{equation}
R_t = \max(0,\, \mathcal{L}_t - \mathcal{L}_{t-1})
\end{equation}

\noindent We set $G_1=R_1=0$. The final importance score $I_t$ for frame $t$ is a
weighted combination of the two signals, each normalized to $[0, 1]$ via min-max
normalization across all $T$ frames. A small constant $\epsilon$ prevents division by
zero when a signal is constant:
\begin{equation}
I_t = \alpha \cdot \frac{G_t - \min_{t} G_t}{\max_{t} G_t - \min_{t} G_t+\epsilon} + 
(1 - \alpha) \cdot \frac{R_t - \min_{t} R_t}{\max_{t} R_t - \min_{t} R_t+\epsilon}
\label{eq:importance}
\end{equation}
where $\alpha = 0.4$ is set empirically to balance the contribution of the two signals.

\subsubsection{Hybrid Downsampling}

At inference, the video is downsampled from $T$ frames to exactly $N_{sel}$ frames. We
first form a Highlight Set $\mathcal{S}$ containing the top-$k$ frames ranked by $I_t$.
We then select a Uniform Set $\mathcal{U}$ of $N_{sel}-k$ frames from the remaining
indices $\{1,\ldots,T\}\setminus\mathcal{S}$. Because the two sets are disjoint, their
sorted union always meets the target budget:
\begin{equation}
\mathcal{F}_{final} = \text{Sort}\!\left(\mathcal{S} \cup \mathcal{U}\right),
\qquad |\mathcal{F}_{final}|=N_{sel}.
\end{equation}
Importance-based selection therefore retains salient changes first, and uniform sampling
fills the remaining slots to preserve temporal coverage. The causal designation applies
to the fast-weight updates; this fixed-budget selection is performed after scores for the
full video are available.
\subsection{System Optimizations for Long-Video Training}
\label{sec:system-optimizations}

Training on long video sequences introduces two memory bottlenecks that standard 
distributed training frameworks do not address: the large activation footprint of 
processing many frames through the ViT backbone, and the combined activation memory 
of the TTT recurrent loop and LLM layers for long packed sequences. The optimizations 
described below directly enable the experimental setting of this paper, extending the 
maximum trainable sequence length by $4\times$ and reducing training step time by up 
to $2.23\times$ at 64 GPUs, making 256-frame training on 40GB GPUs practical without 
sequence parallelism or specialized distributed attention infrastructure.

\subsubsection{Batch-Tiled ViT Processing}

Building on ALST's MLP tiling~\cite{bekman2025arctic}, we partition the ViT frame batch
into independently processed tiles and defer ZeRO-3 gradient reduce-scatter until the
final tile. Frame independence makes this adaptation exact, while replacing repeated
inter-node collectives with one reduce-scatter per layer. It yields speedups from
$1.39\times$ at 8 GPUs to $2.23\times$ at 64 GPUs, where InfiniBand communication is
the dominant bottleneck.

\subsubsection{Asynchronous CPU Activation Offloading}

For long packed sequences, the combined activation memory of the TTT recurrent loop 
and the LLM layers exceeds 40GB even with selective gradient checkpointing. To 
quantify this, for a 28-layer LLM with hidden dimension $D=3{,}584$ and sequence 
length $L=25{,}088$ tokens in 16-bit precision, the activation footprint per layer 
is approximately:
\begin{equation}
\text{Memory}_{\text{layer}} \approx 6 \cdot L \cdot D \cdot 2 \approx 1.08\,\text{GB}
\end{equation}

\noindent where the factor of 6 accounts for the normalization inputs, attention projection 
inputs, and MLP intermediate activations retained under selective checkpointing, 
yielding approximately $30.2$ GB for the full LLM alone, before accounting for ViT 
activations or TTT intermediate states. We address this with an asynchronous CPU offloading pipeline using dedicated CUDA 
streams. Activations are transferred to CPU memory during the forward pass and 
prefetched back to the GPU asynchronously during the backward pass, overlapping 
communication with computation. This extends the maximum trainable sequence length 
by $4\times$ on 40GB hardware. Further implementation details of both optimizations
are provided in Section~\ref{supp:system}. As neither optimization modifies the
attention computation or sequence sharding, they are in principle combinable with 
sequence parallelism or distributed attention techniques for even longer sequences or larger model scales, given sufficient host memory.

\section{Experiments}

\subsection{Experimental Setup}

We evaluate LongVU-TTT on five video understanding benchmarks:
MLVU~\cite{zhou2024mlvucomprehensivebenchmarkmultitask},
LongVideoBench~\cite{wu2024longvideobench}, Video-MME~\cite{videomme},
NExT-QA~\cite{xiao2021nextqanextphasequestionansweringexplaining}, and
LVBench~\cite{wang2025lvbench}. Architecture ablations are conducted on MLVU,
LongVideoBench, and Video-MME using a fixed 30\% subset of the Stage 2 training mixture
shared by all variants to isolate architectural
contributions from data and scale effects.

\subsubsection{Training Setup}

We follow the training protocol of LLaVA-Video~\cite{zhang2024llava}. Training proceeds
in two stages. In the first stage, we train the TTT layer and MLP projector on
LLaVA-CC3M-Pretrain-595K~\cite{sharma2018conceptual} and 200K short-video instruction
samples from
LLaVA-Video-178K~\cite{zhang2024llava}, using a learning rate of
$5 \times 10^{-4}$. In the second stage, all model components are trained jointly on
1.5M samples from LLaVA-Video-178K~\cite{zhang2024llava} and the LLaVA-OneVision image
dataset~\cite{li2024llavaonevisioneasyvisualtask}, with a minimum of 32 and a maximum
of 512 frames per video. We use learning rates of $2 \times 10^{-5}$ for the LLM,
$1 \times 10^{-4}$ for both the MLP projector and TTT layer, and $2 \times 10^{-6}$
for the ViT backbone. Stage 2 uses a global batch size of 128. Batch-tiled ViT
processing and asynchronous activation offloading are enabled throughout Stage 2.

\subsection{Main Results}
\label{sec:main-results}

\begin{table*}[!t]
\centering
\caption{Performance comparison of LongVU-TTT with existing video MLLMs on long and
medium video understanding benchmarks. Frame policy reports either the fixed frame count
or the temporal sampling policy used by each model. LongVU-TTT processes up to 512
frames through TTT-Conv before selecting 128 frames for the LLM. Results from prior work
are shown at their reported precision; our results use two decimals. \textbf{Bold}
denotes the best result and \underline{underline} denotes the second best, including
ties.}
\label{tab:main}
\resizebox{\textwidth}{!}{
\begin{tabular}{lcccccc}
\toprule
\textbf{Model} & \textbf{Frame policy} & \textbf{MLVU} & \textbf{LongVideoBench} & \textbf{Video-MME} & \textbf{NExT-QA} & \textbf{LVBench} \\
\midrule
LongVA~\cite{zhang2024long}         & 128  & 56.3          & -    & 52.6          & -             & -    \\
InternVL2~\cite{InternVL2}          & 16   & 48.1          & 51.8 & 54.0          & -             & -    \\
LongVILA~\cite{chen2024longvila}    & 256  & -             & 57.1 & 60.1          & 80.7          & -    \\
VideoChat2~\cite{li2024mvbench}     & -    & 54.6          & 36.0 & -             & -             & -    \\
LLaVA-OV~\cite{li2024llavaonevisioneasyvisualtask} & 32 & 64.7 & 56.3 & 58.2 & 79.4 & - \\
Qwen2-VL~\cite{wang2024qwen2vlenhancingvisionlanguagemodels} & 256 & 64.2 & 55.6 & \underline{63.3} & - & - \\
Kangaroo~\cite{liu2024kangaroo}     & 64   & 61.0          & 54.8 & 56.0          & -             & 39.4 \\
LLaVA-Video~\cite{zhang2024llava}   & 64   & \underline{70.8} & \underline{58.2} & \underline{63.3} & \underline{83.2} & \underline{42.2} \\
LLaVA-Mini~\cite{zhang2025llava}    & -    & 44.3          & 19.3 & 40.3          & 47.6          & -    \\
LongLLaVA~\cite{wang2024longllavascalingmultimodalllms} & - & 53.3 & 42.1 & 51.6 & 72.2 & - \\
LongVU~\cite{shen2024longvuspatiotemporaladaptivecompression} & 1 fps & 65.4 & 53.5 & 55.3 & 78.0 & - \\
Video-XL~\cite{shu2025video}        & 2048 & 64.9          & 50.7 & 55.5          & 77.5          & -    \\
Vamba~\cite{ren2025vambaunderstandinghourlongvideos} & 1 fps & 65.9 & 55.9 & 57.8 & 78.1 & - \\
\midrule
LongVU-TTT (ours) & $512\!\rightarrow\!128$ & \textbf{71.80} & \textbf{60.40} & \textbf{64.43} & \textbf{83.50} & \textbf{44.30} \\
\bottomrule
\end{tabular}}
\end{table*}

Table~\ref{tab:main} places LongVU-TTT in the broader video-MLLM landscape. This is a
contextual comparison rather than a claim of a new state of the art: models differ in
training data, backbones, and frame policies. Our primary comparison is LLaVA-Video,
which uses the same pipeline, training data, and Qwen2-7B baseline and therefore most
directly isolates the contribution of projector-stage TTT. LongVU-TTT improves over this
reference on all five benchmarks, with the largest gains on LongVideoBench and
LVBench.

\FloatBarrier

\subsection{Controlled Temporal-Operator Study}
\label{sec:operator-study}

\begin{table*}[!t]
\centering
\caption{Controlled temporal-operator study using the same fixed 30\% subset of the
Stage 2 training mixture. All
variants use the same base model, insertion point, feature width, frame input, LLM token
budget, and training setup. (a) compares projector-stage temporal operators; TTT
variants use chunks of 8 frames. (b) isolates fast-weight adaptation, cross-chunk state
carry, and 2D locality.}
\label{tab:operators}
\begin{minipage}[t]{0.495\textwidth}
\centering
\textbf{(a) Temporal operator comparison}\par\smallskip
\resizebox{\linewidth}{!}{%
\begin{tabular}{lccc}
\toprule
\textbf{Method} & \textbf{MLVU} & \textbf{LongVideoBench} & \textbf{Video-MME} \\
\midrule
LLaVA-Video baseline    & 60.35 & 52.54 & 58.20 \\
Bidirectional Mamba2    & 62.46 & 53.55 & 60.80 \\
Gated DeltaNet          & 62.92 & 53.96 & 61.31 \\
TTT-MLP                 & 63.38 & 54.23 & 61.14 \\
TTT-MLP (causal)        & 64.38 & 55.16 & 61.73 \\
Temporal Transformer    & 63.94 & 54.84 & 61.67 \\
TTT-Conv (non-causal)   & 61.52 & 53.46 & 61.08 \\
TTT-Conv (ours)         & \textbf{65.50} & \textbf{56.17} & \textbf{62.11} \\
\bottomrule
\end{tabular}}
\end{minipage}\hfill
\begin{minipage}[t]{0.495\textwidth}
\centering
\textbf{(b) TTT-Conv mechanism ablations}\par\smallskip
\resizebox{\linewidth}{!}{%
\begin{tabular}{lccc}
\toprule
\textbf{Variant} & \textbf{MLVU} & \textbf{LongVideoBench} & \textbf{Video-MME} \\
\midrule
No temporal layer       & 60.35 & 52.54 & 58.20 \\
Static grouped Conv     & 62.11 & 53.48 & 60.12 \\
Reset every frame       & 62.84 & 53.91 & 60.55 \\
Reset every chunk       & 64.21 & 55.08 & 61.44 \\
TTT-Conv, $1\times1$    & 64.36 & 55.02 & 61.52 \\
TTT-Conv, shuffled grid & 63.77 & 54.46 & 60.98 \\
TTT-Conv, $3\times3$    & \textbf{65.50} & \textbf{56.17} & \textbf{62.11} \\
\bottomrule
\end{tabular}}
\end{minipage}
\end{table*}

Table~\ref{tab:operators}(a) compares temporal modules at the same projector location.
All temporal modules improve over independent frames in most settings, while TTT-Conv
is strongest across all three benchmarks. The causal/non-causal Conv comparison isolates
update ordering, and causal TTT-MLP versus TTT-Conv isolates the 2D local fast-weight
operator. Panel (b) shows that the gain is not explained by an added convolution alone:
cross-chunk carry improves all benchmarks, while the $1\times1$ and shuffled-grid
controls reduce the advantage. Detailed variant definitions are in
Section~\ref{supp:architectures}.

\FloatBarrier

\subsection{Temporal Context and Frame Selection}
\label{sec:compression}

\begin{table*}[!t]
\centering
\caption{Temporal contextualization and frame selection. (a) separates the number of
frames processed by the temporal module from the visual budget delivered to the LLM.
(b) compares selection policies with 512 input frames and a fixed 128-frame LLM budget;
all variants use the same TTT-Conv checkpoint.}
\label{tab:context-selection}
\begin{minipage}[t]{0.56\textwidth}
\centering
\textbf{(a) Contextualization and compression}\par\smallskip
\resizebox{\linewidth}{!}{%
\begin{tabular}{lcccc}
\toprule
\textbf{Model} & \textbf{\#Frames} & \textbf{Selection} & \textbf{MLVU} & \textbf{LongVideoBench} \\
\midrule
LLaVA-Video (64 frames)  & 64  & Uniform & 70.80 & 58.20 \\
LLaVA-Video (512 frames) & 128 & Uniform & 65.40 & 55.14 \\
\midrule
LongVU-TTT (512 frames)  & 64  & Uniform & 68.42 & 58.46 \\
LongVU-TTT (512 frames)  & 128 & Hybrid & \textbf{71.80} & 60.40 \\
LongVU-TTT (512 frames)  & 256 & Uniform & 70.80 & 59.42 \\
LongVU-TTT (128 frames)  & 128 & All & 71.50 & \textbf{61.02} \\
\bottomrule
\end{tabular}}
\end{minipage}\hfill
\begin{minipage}[t]{0.425\textwidth}
\centering
\textbf{(b) Frame-selection ablation}\par\smallskip
\resizebox{\linewidth}{!}{%
\begin{tabular}{lcc}
\toprule
\textbf{Selection strategy} & \textbf{MLVU} & \textbf{LongVideoBench} \\
\midrule
Random                  & 69.66 & 58.74 \\
Uniform                 & 70.63 & 59.72 \\
Gradient-direction shift ($G_t$) & 71.19 & 60.07 \\
Alignment-loss delta ($R_t$)     & 71.30 & 60.16 \\
Hybrid ($I_t$)                    & \textbf{71.80} & \textbf{60.40} \\
\bottomrule
\end{tabular}}
\end{minipage}
\end{table*}

Table~\ref{tab:context-selection}(a) shows that 512-frame temporal processing followed
by 128-frame hybrid selection improves over the frame-independent 512-to-128 baseline.
The 64-frame point shows that contextualization also preserves useful information at a
smaller LLM budget, whereas the direct 128-frame result shows that longer
pre-aggregation is not uniformly beneficial. Panel (b) isolates the selector: both
TTT-derived signals improve over uniform sampling, and their
weighted combination $I_t$ is strongest. $G_t$ responds to abrupt changes in update
direction, while $R_t$ captures sustained within-scene novelty. Importance-ranked frames
are selected first and the remaining budget is filled uniformly. The pre-aggregation length sweep remains
in Section~\ref{supp:length}, and qualitative score traces are in
Section~\ref{supp:qualitative}.

\subsection{Temporal Aggregation versus Episodic Retention}
\label{sec:retention}

We identify the temporal evidence window supporting each question in the LongVideoBench
temporal subset and remove that window from the explicit frames passed to the LLM. The
remaining contextualized frames are retained. Questions are grouped by the normalized
gap between the omitted evidence and question timestamp, testing whether evidence can be
recovered from propagated representations rather than directly retained frames.

\begin{table}[H]
\centering
\caption{Evidence-withholding diagnostic on the LongVideoBench temporal subset. Source
evidence is removed from the explicit LLM input for the first three rows; the final row
retains it as an explicit-evidence upper bound. Columns denote normalized temporal-gap
bins between the evidence and question timestamp. \textbf{Bold} denotes the best result
among the evidence-withheld methods.}
\label{tab:retention}
\resizebox{\textwidth}{!}{%
\begin{tabular}{lcccc}
\toprule
\textbf{Method} & \textbf{0--25\%} & \textbf{25--50\%} &
\textbf{50--75\%} & \textbf{75--100\%} \\
\midrule
No temporal layer              & 53.18 & 50.81 & 48.95 & 48.31 \\
TTT-Conv, reset every 8 frames & 54.37 & 51.22 & 49.10 & 48.42 \\
TTT-Conv, state carried        & \textbf{56.42} & \textbf{53.07} & \textbf{49.84} & \textbf{48.66} \\
\midrule
TTT-Conv + explicit evidence (upper bound) & 61.35 & 58.68 & 55.29 & 53.11 \\
\bottomrule
\end{tabular}}
\end{table}

Table~\ref{tab:retention} shows that carrying the fast state improves recovery when
omitted evidence is nearby, with the largest gains at short and medium gaps. The benefit
attenuates beyond half of the video, whereas retaining explicit evidence helps at every
gap. The fast state is therefore a causal temporal aggregator rather than a dependable
episodic store.

\FloatBarrier

\subsection{System Efficiency}

\begin{figure*}[!t]
    \centering
    \includegraphics[width=\textwidth]{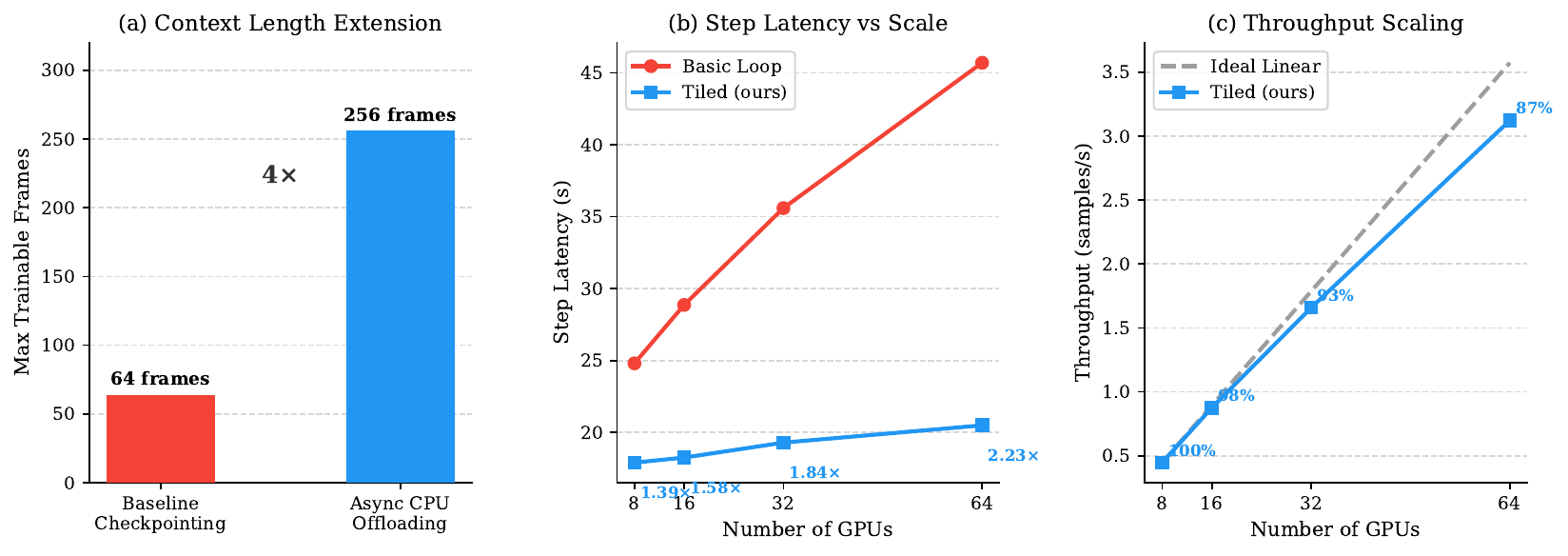}
    \caption{System optimization results measured on A100 40GB GPUs with NVLink
    intra-node and InfiniBand inter-node connections. (a) Maximum trainable frame count
    for standard gradient checkpointing versus
    asynchronous CPU activation offloading on a fixed 8-GPU setup. (b) Training step
    latency for batch-tiled ViT processing versus the standard per-shard loop from 8 to
    64 GPUs at 256 frames with 8 tiles. (c) Throughput scaling efficiency under the same
    256-frame, 8-tile setting relative to ideal linear scaling.}
    \label{fig:system}
\end{figure*}

Figure~\ref{fig:system} validates the optimizations used for long-video training.
Asynchronous CPU activation offloading extends the maximum trainable frame count from 64
to 256 on a fixed 8-GPU setup. Batch-tiled ViT processing defers ZeRO-3 reduce-scatter
until the final tile, yielding speedups from $1.39\times$ at 8 GPUs to $2.23\times$ at
64 GPUs while retaining 87\% throughput-scaling efficiency. Projector-stage latency is
reported separately in Section~\ref{supp:speed}.

\FloatBarrier

\section{Conclusion}

We presented LongVU-TTT, a video MLLM that addresses spatiotemporal redundancy in 
long video understanding through a convolutional Test-Time Training layer inserted 
between the ViT encoder and the LLM. By parameterizing the fast weights as grouped 
convolutions rather than the MLP formulation common in prior TTT work, the layer
uses causal fast-weight updates with explicit 2D local interactions before the LLM. The
gradient signals produced during the inner-loop optimization provide a natural 
importance measure that, combined with uniform sampling, allows compression from 512
to 128 frames while retaining explicit evidence for the LLM. Controlled operator and
retention studies show that causal 2D fast weights are effective temporal aggregators but
do not replace long-horizon episodic storage; their value lies in contextualizing features
before evidence selection. To enable training at this scale on
commodity 40GB hardware, we introduced a batch-tiled ViT processing strategy and 
asynchronous CPU activation offloading that together extend the maximum trainable 
sequence length by $4\times$ and reduce training step time by up to $2.23\times$ at 
64 GPUs without requiring sequence parallelism. LongVU-TTT achieves competitive 
performance across five video understanding benchmarks, improving over its closest
training baseline on all five benchmarks.

\section*{Acknowledgments}

This work was supported by King Abdullah University of Science and Technology
(KAUST). We acknowledge the KAUST Supercomputing Laboratory (KSL) for providing
access to the Ibex cluster for initial runs and GPU experiments. This project was
also granted access to the JURECA supercomputer at the J\"ulich Supercomputing
Centre, Forschungszentrum J\"ulich.

\newpage
\bibliographystyle{iclr2027/iclr2027_conference}
\bibliography{main}
\newpage

\appendix
\section{Supplementary Material}

\subsection{Training Data}
\label{supp:data}

Training proceeds in two stages with different data mixtures designed to progressively 
expose the model to longer and more complex video content.

In Stage 1, we train the TTT layer and MLP projector on LLaVA-CC3M-Pretrain-595K~\cite{sharma2018conceptual} and
200K short-video instruction samples drawn from LLaVA-Video-178K~\cite{zhang2024llava}, capping 
videos at 16 frames. The short-video constraint serves two purposes: it ensures the 
fast weights receive a sufficient number of distinct gradient updates per sample to 
begin learning meaningful temporal representations, and it keeps the memory footprint
manageable for the projector pretraining stage.

In Stage 2, all model components are trained jointly on 1.5M samples drawn from 
LLaVA-Video-178K and the LLaVA-OneVision image dataset~\cite{li2024llavaonevisioneasyvisualtask}. 
Videos are sampled at 1 fps with a maximum of 512 frames, exposing the model to the 
full range of temporal contexts it will encounter at inference. The image samples remain
part of the mixed image-video training distribution.

\subsection{Training Setup}
\label{supp:training}

All experiments are conducted on A100 40GB GPUs with NVLink intra-node and InfiniBand 
inter-node connections. Stage 1 runs on 16 GPUs for 1 epoch with a global batch size 
of 512 and a uniform learning rate of $5 \times 10^{-4}$ for the TTT layer and MLP
projector. Stage 2 scales to 64 GPUs for 1 epoch with a global batch size of 128,
applying component-specific learning rates to reflect the different roles of each 
module: $2 \times 10^{-5}$ for the LLM, $1 \times 10^{-4}$ for both the MLP projector 
and the TTT layer, and $2 \times 10^{-6}$ for the ViT backbone. The higher rate for
the projector components reflects that they are trained from scratch relative to the
pretrained LLM and ViT. Both stages use a cosine decay schedule with a
linear warmup over 3\% of total training steps. Mixed precision training with bfloat16 
is used throughout. Batch-tiled ViT processing with 8 tiles per forward pass and 
asynchronous CPU activation offloading across all LLM layers are both enabled in 
Stage 2, as described in Section~\ref{sec:system-optimizations} of the main paper.

\subsection{System Implementation Details}
\label{supp:system}

For batch-tiled ViT processing, each video frame batch is split into eight tiles. The
tiles are processed independently, while ZeRO-3 gradient reduce-scatter is deferred
until the final tile; concatenating the tile outputs therefore preserves the untiled
forward result and gradient. Parallel causal materialization similarly exposes a
memory--compute tradeoff because it creates a distinct fast-weight state for every frame
in a chunk. We wrap this computation in asynchronous activation checkpointing so these
states are rematerialized during backward rather than retained for the entire step.

For the LLM, activations are offloaded on a dedicated CUDA stream as soon as their
forward consumers finish. During backward, they are prefetched in reverse layer order
and synchronized only before use, overlapping host--device transfers with computation.
Together, checkpointed TTT materialization and CPU offloading bound peak device memory
without changing the temporal update rule or attention computation.

\subsection{Implementation Details}
\label{supp:implementation}

\subsubsection{TTT-Conv Layer}

TTT-Conv uses grouped fast weights with kernel size $K=3$,
$N_g=4$ groups, and an intermediate expansion factor of 4. The chunk size is fixed at 
$C_{size}=8$ frames across all experiments.

Weight initialization is designed to give the TTT update a stable starting point 
while allowing it to gradually activate during training. The down-projection $W_1$ 
uses Kaiming uniform initialization to provide well-conditioned gradients for the 
bottleneck entry, and the up-projection $W_2$ is initialized from a normal distribution 
with standard deviation $1/\sqrt{C}$, where $C$ is the feature width, following 
standard practice for residual projections. The learning-rate projection is initialized
with zero weights and passed through a sigmoid, so every per-group step size begins at
$\operatorname{sigmoid}(0)=0.5$. The momentum projection is initialized with zero weights and a bias of $+3$,
placing initial momentum coefficients close to 1 via a sigmoid. This encourages the model 
to incorporate a large amount of accumulated gradient history from the outset, which we 
found to stabilize the early stages of TTT layer training. The residual connection and 
RMSNorm follow the initialization conventions of the base model.

\subsubsection{Pseudocode}

Algorithm~\ref{alg:ttt} gives the complete forward pass for a single chunk of 
$C_{size}$ frames. The key stages are: parallel gradient computation using the 
chunk-initial weights, application of per-group learning rates and momentum, causal 
prefix-sum materialization of per-frame weight states, per-frame weight normalization, 
and parallel application to queries via \texttt{torch.vmap}. The weight and momentum
states are carried forward to the next chunk to maintain temporal continuity
across the full video. Algorithm indices run from $t=0$ to $C_{size}-1$. The
normalization step is an implementation stabilization omitted
from the compact recurrence in the main paper: before application, each materialized
kernel is rescaled to its target norm $\eta_j$, and the same normalization is applied to
the state carried into the next chunk.

\begin{algorithm}[H]
\caption{TTT-Conv Forward Pass (Single Chunk)}
\label{alg:ttt}
\begin{algorithmic}[1]
\Require Frame features $X_c \in \mathbb{R}^{C_{size} \times C \times H \times W}$,
         fast weights $(W_1^{init}, W_2^{init})$,
         momentum state $(p_1^{init}, p_2^{init})$,
         weight norms $(\eta_1, \eta_2)$
\Ensure Enriched features $\hat{X}_c$, updated weights and momentum state

\State Compute projections: $k, v, q \leftarrow \text{Linear}(X_c)$; reshape to $(C_{size}, C, H, W)$
\State $lr \in \mathbb{R}^{C_{size} \times N_g \times 2} \leftarrow \operatorname{sigmoid}(\text{Linear}(X_c))$, averaged over spatial dims
\State $m \in \mathbb{R}^{C_{size} \times N_g} \leftarrow \operatorname{sigmoid}(\text{Linear}(X_c) + 3)$, averaged over spatial dims

\Statex \textbf{// Step 1: Parallel forward pass using chunk-initial weights}
\State $z_t \leftarrow \sigma(\text{Conv2d}(k_t, W_1^{init}))$ for all $t$ \Comment{all frames in parallel}

\Statex \textbf{// Step 2: Parallel gradient computation (batched einsum)}
\State $g_j^{(t)} \leftarrow -\nabla_{W_j}\mathcal{L}(k_t,v_t;W_1^{init},W_2^{init})$
\Statex \hfill for $j\in\{1,2\}$ and all $t$

\Statex \textbf{// Step 3: Scale by per-group learning rates}
\State $u_j^{(t)} \leftarrow lr_t^{(j)} \cdot g_j^{(t)}$ \hfill for $j \in \{1,2\}$, all $t$

\Statex \textbf{// Step 4: Add momentum contribution (frame $t$ sees history up to $t-1$)}
\State $h_j^{(t)} \leftarrow p_j^{init} + \sum_{i=0}^{t-1} u_j^{(i)}$ \Comment{cross-chunk history via $p_j^{init}$}
\State $\delta_j^{(t)} \leftarrow u_j^{(t)} + m_t \cdot h_j^{(t)}$ \hfill for $j \in \{1,2\}$

\Statex \textbf{// Step 5: Causal prefix-sum — frame $t$ excludes its own update}
\State $W_j^{(t)} \leftarrow W_j^{init} + \sum_{i=0}^{t-1} \delta_j^{(i)}$ \hfill for all $t$

\Statex \textbf{// Step 6: Per-frame weight normalization}
\State $W_j^{(t)} \leftarrow W_j^{(t)} \cdot \eta_j\, /\, \|W_j^{(t)}\|$ \hfill for all $t$

\Statex \textbf{// Step 7: Parallel apply to queries (torch.vmap)}
\State $\hat{x}_t \leftarrow \text{Conv2d}\!\left(\sigma\!\left(\text{Conv2d}(q_t, W_1^{(t)})\right),\, W_2^{(t)}\right)$ for all $t$
\State $\hat{X}_c \leftarrow (\hat{x}_1, \ldots, \hat{x}_{C_{size}})$ reshaped to $(C_{size}, C, H \cdot W)$

\Statex \textbf{// Step 8: Carry state to next chunk}
\State $W_j^{final} \leftarrow \text{Normalize}\!\left(W_j^{init} + \textstyle\sum_t \delta_j^{(t)},\; \eta_j\right)$
\State $p_j^{next} \leftarrow \delta_j^{(C_{size}-1)}$ \Comment{momentum state carried to the next chunk}
\State \Return $\hat{X}_c$,\; $(W_1^{final}, W_2^{final})$,\; $(p_1^{next}, p_2^{next})$
\end{algorithmic}
\end{algorithm}

\subsection{Ablation Architecture Details}
\label{supp:architectures}

All variants compared in Table~\ref{tab:operators} of the main paper are inserted at the same position 
in the pipeline, between the ViT encoder and the spatial pooling layer, and trained 
under identical conditions using the same fixed 30\% subset of the Stage 2 training
mixture. All chunked variants use 
a chunk size of $C_{size}=8$ frames. The descriptions below focus on the specific 
design decisions that distinguish each variant from TTT-Conv, to make the controlled 
nature of the comparison transparent.

\subsubsection{Bidirectional Mamba2}

The bidirectional Mamba2 variant follows the VideoMamba~\cite{li2024videomamba} design, 
applying two parallel selective state-space scans over the token sequence, one in the 
forward temporal direction and one in reverse, and summing their outputs. We use a 
state size of 288. Frame features are flattened from $H \times W \times C$ to
$(H \cdot W) \times C$ before the scan and reshaped back afterward; the scan does not
explicitly impose 2D local-neighborhood interactions. Unlike the TTT variants, Mamba2 processes the full 
token sequence in a single pass and requires no chunking.

\subsubsection{Gated DeltaNet}

We use the Gated DeltaNet layer from the Flash Linear Attention library
(FLA)~\cite{yang2024fla}, which implements the gated delta-rule recurrence of
\citet{yang2024gated}. It is inserted after the ViT with the same feature
width and output interface as the other variants, uses a state size matched to the
Mamba2 parameter budget, and scans frames causally before spatial pooling.

\subsubsection{Temporal Transformer}

The temporal Transformer is a pre-norm residual attention block implemented with
FlexAttention. At each spatial location, it applies attention across the frame dimension
through a custom causal mask, followed by its MLP sublayer. It uses the same hidden width
and projector location as TTT-Conv. This temporal operator has quadratic cost in the
number of frames.

\subsubsection{TTT-MLP}

The TTT-MLP variant adopts LaCT's hardware-efficient MLP fast-weight
formulation~\cite{zhang2025test}, using an expansion factor of 4 and 4 heads. It does
not use LaCT's sliding-window attention because local information is already modeled by
the ViT. The key distinction from TTT-Conv is the treatment of each chunk: all frames
within a chunk are processed
as a single joint update and apply operation, with no intra-chunk causal ordering. The 
weight state is updated once per chunk from the aggregated gradient across all frames, 
and the same updated weights are applied to all queries in that chunk. There is no 
prefix-sum materialization of intermediate per-frame states. The same negative dot 
product alignment objective is used as in TTT-Conv, and the flattened operator does not
explicitly impose 2D local-neighborhood interactions.

\subsubsection{Causal TTT-MLP}

The causal TTT-MLP differs from TTT-MLP only by materializing a distinct prefix-update
state for each frame, following the causal rule used by TTT-Conv.

\subsubsection{TTT-Conv Non-Causal}

The non-causal TTT-Conv keeps the TTT-Conv parameterization but replaces per-frame
prefix states with one update aggregated over each chunk and shared by all queries in
that chunk. All other components and initialization are unchanged.

\subsection{Pre-Aggregation Context Length}
\label{supp:length}

We vary the number of frames processed before compression while fixing the LLM budget at
128 frames. Table~\ref{tab:length} shows that increasing pre-aggregation context beyond
128 frames yields only small MLVU changes and is non-monotonic on LongVideoBench. This
behavior motivates our distinction between temporal contextualization and persistent
episodic storage.

\begin{table}[t]
\centering
\caption{Input-length sweep at a fixed 128-frame LLM budget.}
\label{tab:length}
\begin{tabular}{lcc}
\toprule
\textbf{\#Frames} & \textbf{MLVU} & \textbf{LongVideoBench} \\
\midrule
128 & 71.50 & \textbf{61.02} \\
256 & 71.72 & 60.86 \\
512 & \textbf{71.80} & 60.40 \\
\bottomrule
\end{tabular}
\end{table}

\subsection{Inference Speed Comparison}
\label{supp:speed}

A natural question is whether convolutional fast weights incur meaningful overhead.
Table~\ref{tab:speed} compares isolated projector-stage latency on one A100 40GB GPU at
identical feature dimensions; TTT variants use $C_{size}=8$.

\begin{table*}[t]
\centering
\caption{Temporal-module inference latency (ms) on one A100 40GB GPU.}
\label{tab:speed}
\resizebox{\textwidth}{!}{%
\begin{tabular}{lcccc}
\toprule
\textbf{\#Frames} & \textbf{TTT-Conv} & \textbf{TTT-MLP} &
\textbf{Gated DeltaNet} &
\textbf{\shortstack{Temporal\\Transformer}} \\
\midrule
64   & 123.97  & 26.98  & 13.83  & 217.85 \\
128  & 247.99  & 45.82  & 27.37  & 862.42 \\
256  & 496.60  & 91.76  & 54.23  & 3448.04 \\
512  & 993.15  & 184.60 & 108.89 & 13704.77 \\
1024 & 1987.31 & 369.38 & 219.25 & 54921.84 \\
\bottomrule
\end{tabular}}
\end{table*}

The recurrent fast-weight operators scale approximately linearly over this range, whereas
the temporal Transformer exhibits the expected quadratic scaling with the number of
frames. Its frame-wise temporal attention is implemented with FlexAttention as described
in Section~\ref{supp:architectures}. TTT-Conv is slower than TTT-MLP and Gated DeltaNet
because it applies grouped 2D fast weights, but remains substantially faster than the
temporal Transformer at every tested context length. This temporal enrichment occurs once
before the more expensive LLM stage.

\subsection{Additional Qualitative Results}
\label{supp:qualitative}

\begin{figure*}[t]
    \centering
    \includegraphics[width=\textwidth]{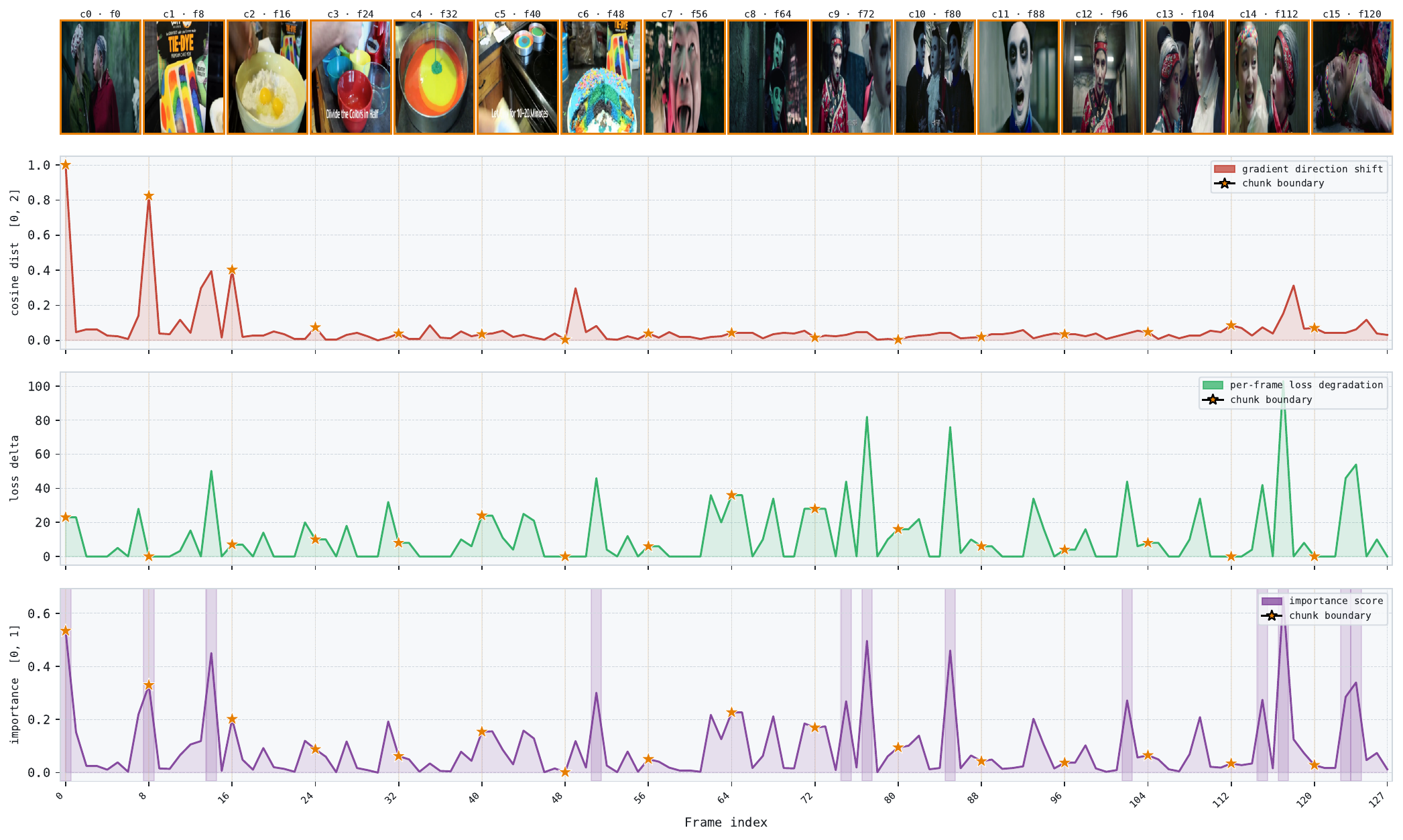}
    \caption{Importance score components across 128 frames of a representative video.
    From top to bottom: gradient-direction shift $G_t$, alignment-loss delta $R_t$, and
    combined importance $I_t$ from Equation~\ref{eq:importance}. Star markers indicate
    chunk boundaries, and shaded regions indicate selected frames.}
    \label{fig:importance}
\end{figure*}

Figure~\ref{fig:importance} visualizes the two importance signals on a representative
video. The gradient-direction shift $G_t$ responds most strongly to abrupt scene-level
transitions, producing isolated peaks where the fast-weight update direction changes.
The alignment-loss delta $R_t$ instead captures sustained within-scene changes. Their
combination selects both sudden transitions and gradual but meaningful visual changes.

\begin{figure*}[t]
    \centering
    \includegraphics[width=\textwidth]{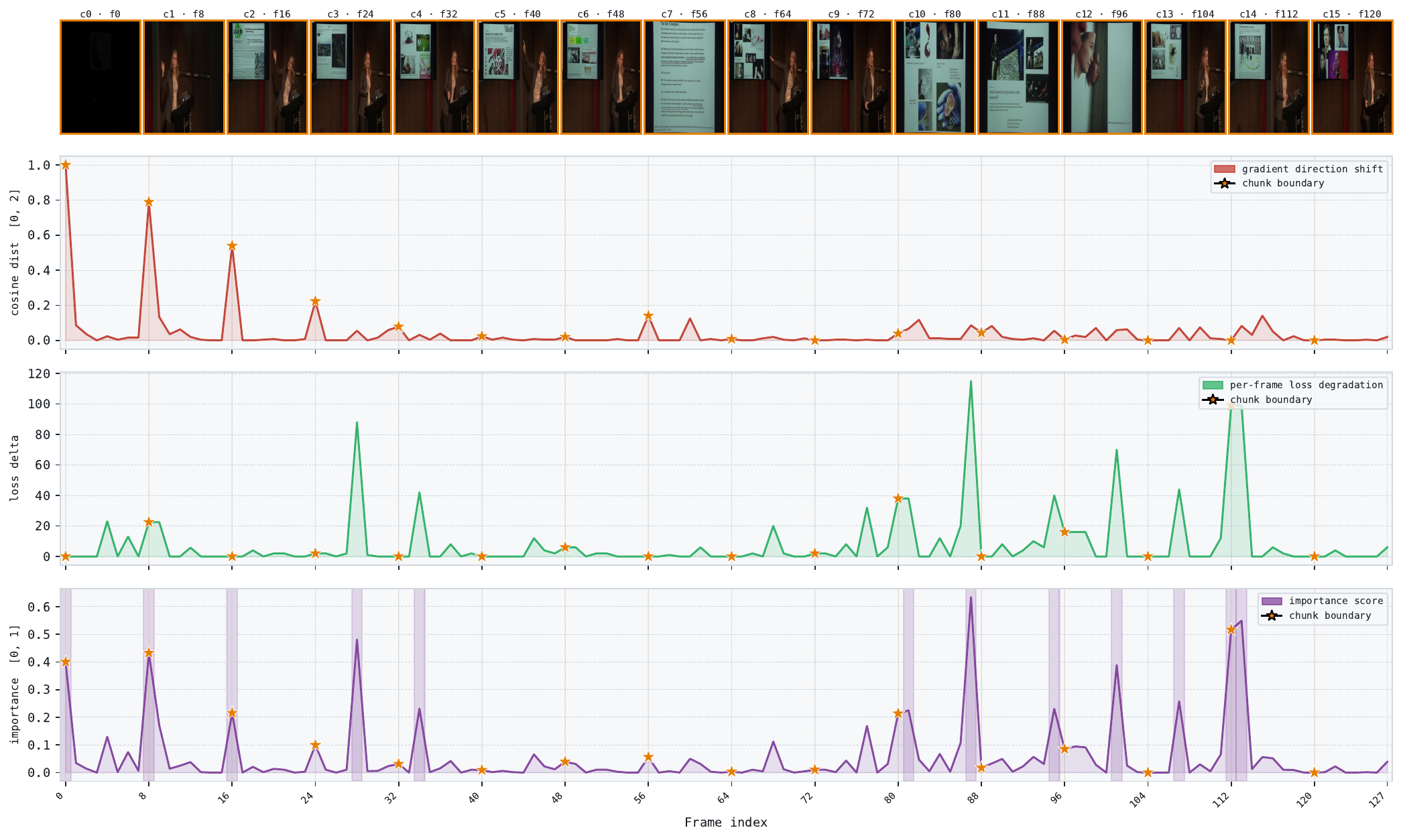}
    \caption{Importance score components $G_t$, $R_t$, and $I_t$ for a lecture video 
    from LongVideoBench~\cite{wu2024longvideobench}. Star markers indicate chunk 
    boundaries and shaded regions indicate selected frames.}
    \label{fig:supp_qual}
\end{figure*}

Figure~\ref{fig:supp_qual} shows a representative example from a long lecture video 
drawn from LongVideoBench~\cite{wu2024longvideobench}, in which a speaker presents 
in front of a series of academic posters. This setting is particularly challenging 
for frame selection because the visual content changes subtly and gradually rather 
than through abrupt cuts. The gradient-direction shift $G_t$ fires most strongly in 
the early frames, capturing the initial scene establishment and early camera 
adjustments, but remains largely flat for the remainder of the video where the scene 
layout is stable. In contrast, the alignment-loss delta $R_t$ provides a richer 
and more informative signal throughout, with pronounced peaks distributed across the 
full video that correspond to moments where new poster content, gestures, or visual 
elements enter the frame. The combined importance score $I_t$ selects frames that 
would be missed by either signal alone: the early transitional frames identified by 
$G_t$ and the later content-rich moments identified by $R_t$. This example illustrates 
that for videos dominated by subtle within-scene variation rather than hard cuts, the 
alignment-loss delta is the more informative of the two signals, and the hybrid 
combination is essential for adequate temporal coverage.

\end{document}